\documentclass[11pt]{article}

\usepackage{acl}

\usepackage{times}
\usepackage{latexsym}

\usepackage[T1]{fontenc}

\usepackage[utf8]{inputenc}

\usepackage{microtype}

\usepackage{inconsolata}

\usepackage{graphicx}
\usepackage{amsmath}
\usepackage{amssymb}
\usepackage{booktabs}
\usepackage{multirow}
\usepackage{float}
\usepackage{placeins}
\usepackage{stfloats}
\usepackage{adjustbox}
\usepackage[nameinlink,noabbrev]{cleveref}
\usepackage{soul}
\usepackage{color, xcolor}
\definecolor{lightblue}{RGB}{0, 191, 255}
\sethlcolor{lightblue}
\soulregister\cite7
\soulregister\eqref7
\soulregister\ref7

\title{A Multi-cluster Boundary Learning Method for Out-of-Scope Intent Detection via MiniLM Embedding}

 \author{
   \textbf{Yihong Xu\textsuperscript{1}},
   \textbf{Mingyu Kang\textsuperscript{1}\thanks{Corresponding authors.}},
   \textbf{Linyuan L\"u\textsuperscript{1}\footnotemark[1]}\\
   \textsuperscript{1}University of Science and Technology of China, Hefei, China\\
   \texttt{xuyihong@mail.ustc.edu.cn}\\
   \texttt{\{kangmingyu, linyuan.lv\}@ustc.edu.cn}
 }

\begin{document}
\maketitle

\begin{abstract}
Intent detection is a critical task that bridges human intents and system actions in human-machine interaction systems. However, there still exist challenges for detecting out-of-scope (OOS) intents. (i) The traditional methods view the OOS intent detection as a multi-class classification, then the detection accuracy decreases as the class number of the known intents increases; (ii) LLM-embedding methods require large parameters, that makes them difficult to train and practically deploy. Thus, this work proposes a multi-cluster boundary learning method to detect OOS intents via MiniLM embedding (i.e., \texttt{all-MiniLM-L6-v2}) in an one-class classification workflow. The method learns the boundaries of multi-cluster embeddings generated by MiniLM from the training utterances, and then rejects the out-of-domain utterances as OOS intents. Experiments are conducted on public CLINC150, StackOverflow and Banking77 datasets. The results show that the method achieves the state-of-the-art OOS intent detection performance compared the other baselines. Ablation studies are also conducted and the results show that the used MiniLM can better adapt to the workflow and utterance embedding requirements. The code is available at supplementary materials.
\end{abstract}

\section{Introduction}
\begin{figure}[t]
    \centering
    \includegraphics[width=\columnwidth]{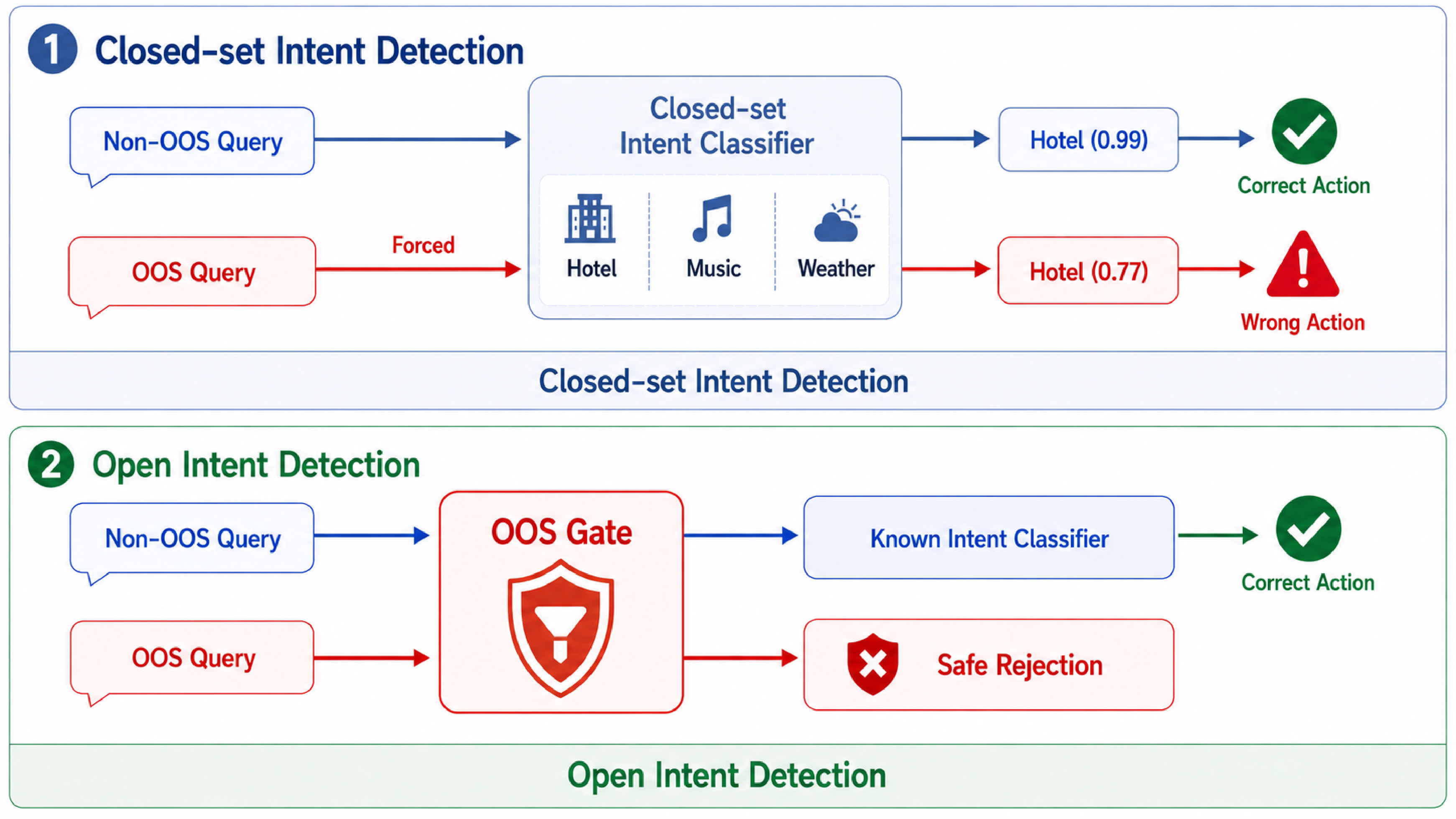}
    \caption{A diagram of OOS intent detection}
    \label{fig:motivation}  
\end{figure}

Intent detection is a critical yet challenging task for human-machine interaction systems. It maps user utterances to system actions, which bridges human intents and system behaviors~\citep{tur2010what,louvan2020recent,wolflein2025agents}. However, human intents are complex and cannot be fully enumerated in a closed-set intent detection system. That means if the system receives an out-of-scope (OOS) intent utterance but fails to reject it, that will trigger incorrect actions and induce harmful consequences, as shown in figure~\ref{fig:motivation}. Thus, a more important and difficult task is to detect the OOS intents and reject them in a timely manner, which is actually an open intent detection task~\citep{hoffman2024inferring, muzahid2024survey}.

There are three types of intent detection methods, including statistical methods, deep-learning methods and LLM-embedding methods. Among them, the statistical methods formulate intent detection as a supervised text classification problem. These methods combine manually designed lexical features with traditional statistical classifiers, such as support vector machine~\citep{haffner2003optimizing} and naive Bayes classifier~\citep{schuurmans2020intent}. But the deep-learning methods use deep neural networks, e.g., convolutional neural network (CNN)~\citep{kim2014convolutional}, recurrent neural network (RNN)~\citep{liu2016attention} and transformer network~\citep{vaswani2017attention}, to perform intent detection via semantic feature extraction. Early deep-learning classifiers are trained on a closed-set mode, and then identify unknown out-of-domain utterances as known intents~\citep{hendrycks2017baseline, guo2017calibration}. Thus, some types of open intent detection are further proposed to conduct safe rejection for the unknown utterances with OOS intents. They are class-wise open classification method~\citep{shu2017doc}, distance-based scoring method~\citep{lee2018simple}, contrastive representation learning method~\citep{zeng2021modeling} and adaptive boundary learning method~\citep{zhang2021adaptive}. After that, with the strong semantic representation ability of LLMs, the LLM-embedding methods are proposed to extract utterance features by semantic embedding, prompt-based reasoning and uncertainty-aware agent routing~\citep{arora2024intent}.

However, the existing methods still face two limitations. First, the OOS queries with similar semantic meaning as known intents are easily absorbed into the decision regions of the known intents~\citep{li2025multi}, but in fact they are OOS. Thus, if mix OOS labels into known intent labels, and simply view the OOS intent detection as a multi-class classification, the accuracy of OOS rejection would decrease as the class number increases. Second, although LLM-embedding methods improve the ability of semantic understanding, they still require very large parameters, that induces remarkable computational costs and high sensitivity of prompt designs~\citep{zaera2025efficient}. Thus, they are difficult to train and deploy in real-time resource-constrained dialogue systems.

To address these limitations, this work proposes a multi-cluster boundary learning method for OOS intent detection with a cascade workflow. The workflow decouples the complex open intent detection problem into multiple simpler stages. This allows the workflow to use a lightweight MiniLM instead of LLM wih large-scale parameters. Moreover, it reduces the OOS intent detection to a one-class classification, instead of a complex multi-class classification. Moreover, we found that MiniLM can embed the observation data into multiple clusters. Thus, if the boundaries of clusters are learned, the out-of-domain utterances can be identified as OOS intents.

Thus, the main contributions are as follows:
\begin{enumerate}
    \item[1.] A multi-cluster boundary learning method is proposed. This method learns the boundaries of multi-cluster embeddings generated by MiniLM from the training utterances, and then rejects the out-of-domain utterances as OOS intents. 
    
    \item[2.] A cascade workflow is proposed to first conduct OOS intent detection as an one-class classification, and then conduct known intent detection as a close-set multi-class classification. This means the OOS intent detection task can be addressed independently. 

    \item[3] Experiments are conducted on public real-world datasets for OOS intent detection task. The results show that the method achieves the state-of-the-art performance for OOS intent detection. Moreover, ablation studies are also conducted, and the results show that the used MiniLM, \texttt{all-MiniLM-L6-v2}, can better adapt to the workflow and utterance embedding requirements.  
\end{enumerate}

\section{Related Work}
The open intent detection methods can be mainly categorized into three types, i.e., statistical methods, deep-learning methods and LLM-embedding methods, according to the problem settings and the types of classifiers.

\subsection{Statistical Methods for Intent Detection}
\label{sec: cascade}
The statistical methods use hand-crafted lexical features with statistical classifiers to detect the intent classes. Among them, \citet{wang2002combination} proposes a method combining rule-based grammars with statistical classifiers, but the grammar must be hand-crafted for each domain. To reduce the dependence, \citet{haffner2003optimizing} use support vector machine as discriminative model, which shows strong performance on intent classification without rule-based components. But these methods still rely on manually designed features. Thus, \citet{tur2010what} proposes an n-gram method to model local contextual dependencies, but feature sparsity limits its effectiveness on diverse intent expressions. Then, \citet{schuurmans2020intent} compare these methods, and show that dense word embeddings outperform the sparse features for intent classification.

\subsection{Deep-Learning Methods for Intent Detection}
Then, the deep-learning methods use artificial neural networks to parametrize intent classifiers and learn dense semantic representations. E.g., CNN is used to extract the local semantic patterns from word embeddings~\citep{kim2014convolutional}, and RNN is used to capture the sequential dependencies for joint intent detection and slot filling~\citep{liu2016attention}. Moreover, Transformer model is used to directly capture the global semantic dependencies through self-attention mechanism~\citep{vaswani2017attention}. These methods improve the in-distribution accuracy, but the classification mode is still closed-set, and the classifier can only identify unknown utterances with known intent labels, and cannot reject the OOS intents~\citep{hendrycks2017baseline, guo2017calibration}. 

Thus, to reject the OOS intents, there are several methods, that view the OOS intent detection as a open-set classification task. E.g., \citet{hendrycks2017baseline} set an extra OOS class, and transform the open-set classification into a closed-set one. Then, the method assigns regular scores to each class through softmax probability. Also, \citet{lee2018simple} propose a Mahalanobis-based scoring method, that models the score as class-conditional Gaussian distribution. But these score assumptions usually violate the real-world class distributions. Moreover, with the increasing class number, the OOS labels couple with many known intent labels, then the score of OOS class is diluted and the accuracy decreases. To improve the representation separability, \citet{zeng2021modeling} propose a supervised contrastive learning, that learns discriminative embeddings to better separate intent categories. But this still yields limited improvements. 

Then, \citet{shu2017doc} drop the transformation and propose a open classification method. The method replaces the softmax probability with one-vs-rest sigmoid classifiers, and then calculates a class-wise rejection threshold. If the softmax score is over the threshold, the input utterance would be rejected. Then, \citet{zhang2021adaptive} propose an decision boundary learning method to learn a boundary from all embedding known intents, and the external region is OOS. But the method can only learn a single centroid for each intent. Actually, the embeddings usually have multiple centroids based on our observation. Then, \citet{li2025multi} propose a multi-granularity boundary learning method to learning multiple boundaries for each intent.

\subsection{LLM-Embedding Methods for Intent Detection}
Then, with the LLM technique, the LLM-embedding methods are proposed to strengthen the semantic representations and feature embeddings for user utterance by multi-agent routing mode. Among them, \citet{arora2024intent} proposes a hybrid method, that uses a Sentence Transformer to route uncertain intent predictions to a designated LLM. But this LLM is only designed for uncertain queries and is not trained for OOS queries. Then, \citet{zaera2025efficient} propose an uncertainty-aware routing method, that only triggers a fine-tuned LLM according to the Monte Carlo dropout criterion. But the routing decisions are still unstable if there exists distribution drift. And, the multi-agent LLMs require large computation cost, that make it difficult to deploy for real-world applications. Moreover, \citet{chen2025collaborating} propose a small-large model collaboration method for few-shot intent detection. The method uses LLM to generate OOS utterance and conduct data augmentation during pretraining. And then, it uses the smaller deep-learning models to realize intent detection. 

Thus, the existing methods still face two problems. (i) the OOS intent detection is usually viewed as a multi-class classification and hardly separated from the known intent detection. (ii) The multi-agent LLMs require remarkably large computation cost. And due to the reason (i), the current routing mechanism is not accurate.

\section{Method}
\subsection{Cascade Workflow for Intent Detection}
\begin{figure*}[t]
    \centering
    \includegraphics[width=\textwidth]{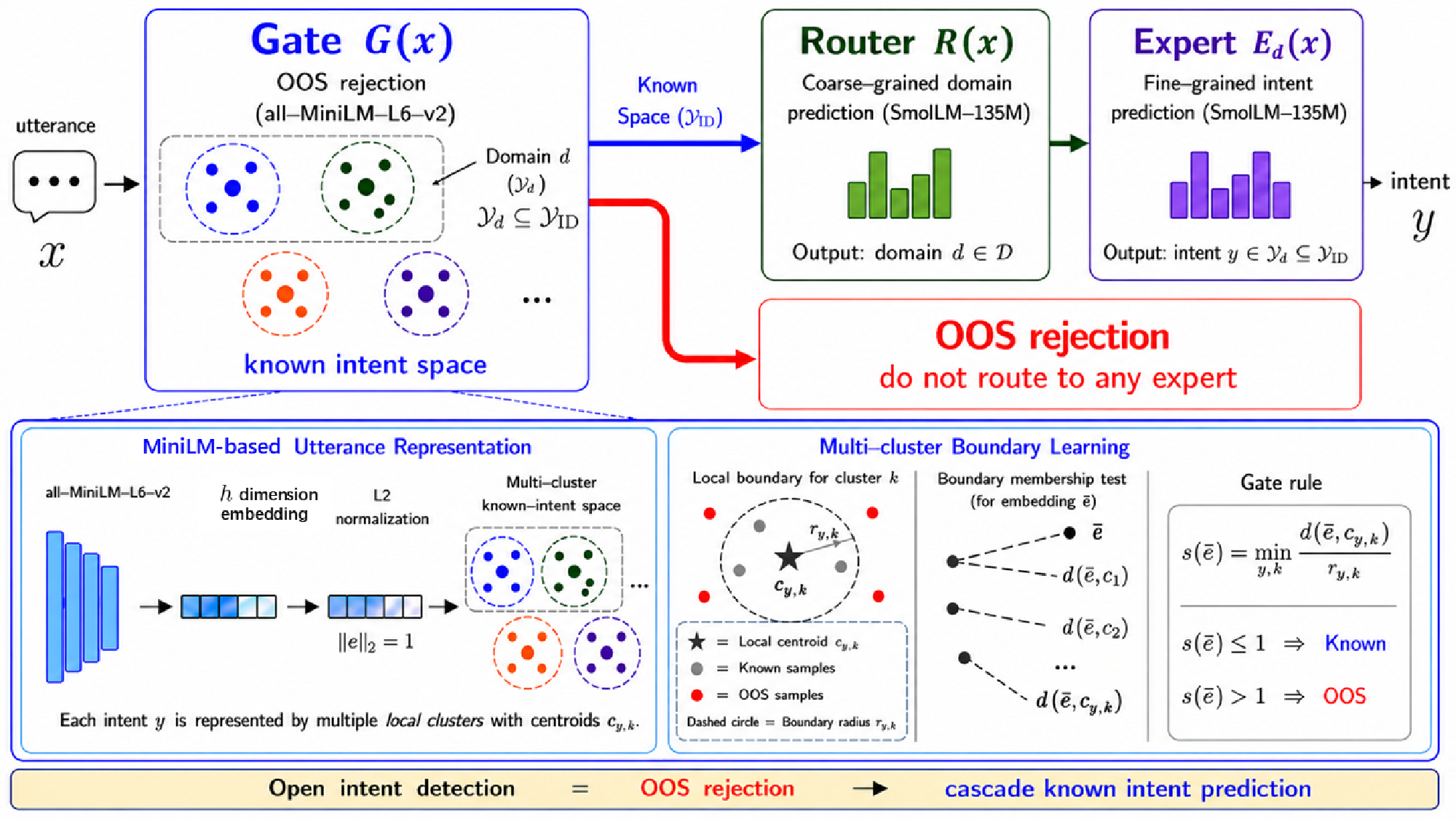}
    \caption{An Overview of the cascaded workflow via MiniLM-based multi-cluster boundary learning.}
    \label{fig:model_architecture}
\end{figure*}

The proposed method uses a three-stage cascade: gate $\rightarrow$ router $\rightarrow$ expert, as shown in figure~\ref{fig:model_architecture}. Given an input utterance $x$, the gate module first determines the OOS rejection. If $x$ is rejected, the gate module outputs an OOS label, otherwise, $x$ is inputted into the following modules. Then, the router module discriminates the coarse-grained class domain for the remained known intents. Then the utterance $x$ is delivered to an expert module w.r.t the class domain for fine-grained discrimination of a few classes. Note that, due to the fact that the known intent classification is close-set, all intents for training is known, thus the router and experts are all parametrized by SmolLM-135M, that is supervisedly fine-tuned through low-rank adaptation (LoRA)~\citep{hu2022lora} technique on the observation data of known intents.

For an input utterance $x$, the gate $G(\cdot)$ first checks whether $x$ is OOS:
\begin{equation}
z = G(x), \quad z \in \{\mathrm{ID}, \mathrm{OOS}\},
\end{equation}
where $z$ is the binary decision, $\mathrm{ID}$ means the input is in-distribution, and $\mathrm{OOS}$ means the input is out-of-scope. If $z = \mathrm{OOS}$, an OOS label is returned. Otherwise, the router $R(\cdot)$ predicts the domain:
\begin{equation}
d = R(x), \quad d \in \mathcal{D},
\end{equation}
where $\mathcal{D}$ is the domain set. Then the expert $E_d(\cdot)$ classifies the intent:
\begin{equation}
y = E_d(x), \quad y \in \mathcal{Y}_d \subseteq \mathcal{Y}_{\mathrm{ID}},
\end{equation}
where $\mathcal{Y}_{\mathrm{ID}}$ is the known intent label set and $\mathcal{Y}_d$ is the intent subset for domain $d$.

The proposed cascade workflow separates OOS detection from known-intent classification. Each module handles a simpler sub-task to reduces the complexity. As a result, the Gate only needs to determine whether an input belongs to the known-intent space. Thus, a lightweight model is sufficient.

\subsection{MiniLM-Based Intent Embedding}
Since the gate only needs to distinguish OOS from known-intent samples, a lightweight MiniLM encoder, \texttt{all-MiniLM-L6-v2}, is used to construct the semantic representation space. The MiniLM maps each input utterance $x$ to an embedding with fixed dimension:
\begin{equation}
e = f_{\theta}(x),\quad e \in \mathbb{R}^{h},
\label{eq: embed}
\end{equation}
where $f_{\theta}(\cdot)$ denotes the MiniLM-based encoder parameterized by $\theta$, and $h$ denotes the embedding dimension. Note that, the MiniLM with parameter $\theta$ has no need to train in this work, that implicitly assumes the MiniLM can directly embed the utterance into multiple clusters, and in fact, that is indeed the case through empirical experiments. Then, the gate applies $L_2$ normalization before computing distances to the centroids:
\begin{equation}
\bar{e} = \frac{e}{\|e\|_2}.
\end{equation}
The normalized embedding $\bar{e}$ is then used to construct the known-intent geometry for OOS detection.
    
The embedding space is not used solely as a generic feature space. It also defines the local structure of known intents.But a single-centroid assumption can be too restrictive.Because utterances with the same intent label may differ in wording, expression, and semantic focus. Thus, forcing a known intent into a single compact region can produce an inaccurate boundary. This increases the risk of rejecting valid in-domain samples or accepting OOS samples. Thus, each known intent is represented by multiple local clusters, and each cluster has a centroid.

For intent $y$, its centroid set is defined as:
\begin{equation}
\mathcal{C}_{y}=\{c_{y,1}, c_{y,2}, \ldots, c_{y,K_y}\},
\label{eq: ky}
\end{equation}
where $c_{y,k}, k=1, \dots, K_y$, denotes the $k$-th centroid of known intent $y$, and $K_y$ is the number of centroids. The centroids are constructed via K-means clustering within each intent. The number of centroids is controlled by a global setting and can be overridden for specific intents. If the number of known intents is $M$ and each intent has $K$ centroids, the gate maintains $M \times K$ centroids in total. These centroids are local geometric representatives of the known intents. Thus the representation provides the basis for boundary learning.

\subsection{Multi-cluster Boundary Learning for OOS Intent Detection}
The multi-cluster representation defines local semantic regions for known intents, and OOS detection is therefore formulated as a boundary learning problem in the embedding space. An input is accepted if it falls inside at least one learned known-intent cluster region and is rejected if it lies outside all known-intent regions. To achieve this, the gate learns a local boundary around the centroid of each cluster to separate supported inputs from OOS inputs.

For the centroid of each cluster $c_{y,k}$, the gate estimates a local radius $r_{y,k}$ from the training samples assigned to that specific cluster. In the default configuration, the radius is computed as follows:
\begin{equation}
r_{y,k}=\mu_{y,k} + \lambda \sigma_{y,k},
\label{eq: lambda}
\end{equation}
where $\mu_{y,k}$ and $\sigma_{y,k}$ are the mean and standard deviation of the distances from the assigned samples to $c_{y,k}$, and $\lambda$ controls the boundary width.

The distance function is a diagonal Mahalanobis distance. For an embedding $\bar{e}$ and a centroid $c$, the distance is computed as:
\begin{equation}
d(\bar{e},c)=\sqrt{\sum_j(\bar{e}_j-c_j)^2\frac{1}{\sigma_j^2+\epsilon}},
\end{equation}
where $\bar{e}_j$ is the $j$-th component of $\bar{e}$ and $c_j$ is the same componenet of $c$. Moreover, $\sigma_j^2$ is the corresponding feature-wise variance estimated from samples assigned to the corresponding cluster, and $\epsilon$ is a small regularization constant. This distance weights each embedding dimension by its estimated variance, which reduces the influence of dimensions with large natural variation.

For an incoming utterance, the gate compares its normalized embedding with all known-intent centroids and computes the normalized nearest-boundary score:
\begin{equation}
s(\bar{e})=\min_{y \in \mathcal{Y}_{\mathrm{ID}},\; 1 \le k \le K_y}\frac{d(\bar{e},c_{y,k})}{r_{y,k}},
\label{eq: ryk}
\end{equation}
where $c_{y,k}$ denotes the $k$-th local centroid of known intent $y$, and $r_{y,k}$ denotes the corresponding local boundary radius. The ratio of Eq.~(\ref{eq: ryk}) measures the distance from $\bar{e}$ to this centroid relative to the learned local boundary size.

The gate decision is:
\begin{equation}
G(x)=
\begin{cases}
\mathrm{ID}, & s(\bar{e}) \le 1 \\
\mathrm{OOS}, & s(\bar{e}) > 1
\end{cases}
\end{equation}
The threshold is one because each distance is normalized by its radius. If $s(\bar{e}) \le 1$, the input falls inside at least one known-intent region and is accepted. If $s(\bar{e}) > 1$, the input lies outside all known-intent regions and is rejected as OOS. The gate only decides whether an input belongs to the known intent space. It does not assign a specific intent label.

\section{Experiments}
\subsection{Datasets}
Three datasets are used for performance evaluation, as shown in Table~\ref{tab:dataset_stats}. For each dataset, they are split into training, validation and test sets under 6:1:3.

\textbf{CLINC150} is a large-scale multi-domain dataset with 150 in-scope intents and explicit OOS samples~\citep{larson2019evaluation}. The data source is available at \url{https://github.com/clinc/oos-eval}.

\textbf{StackOverflow} is a technical-domain short-text dataset with 20 intent classes and substantial semantic overlap across classes. The data source is available at \url{https://github.com/jacoxu/StackOverflow}.

\textbf{Banking77} is a fine-grained, single-domain dataset with 77 intent classes. It focuses on high-granularity classification in the financial domain. The data source is available at \url{https://github.com/PolyAI-LDN/task-specific-datasets}.

\begin{table}[H]
\centering
\renewcommand{\arraystretch}{1.05}
\resizebox{\columnwidth}{!}{
\begin{tabular}{lccc}
\toprule
\textbf{Dataset} & \textbf{\#Intents} & \textbf{\#Samples} & \textbf{Type} \\
\midrule
CLINC150       & 150              & 22500 & Multiple domains \\
StackOverflow  & 20                     & 20000 & Technical \\
Banking77  & 77 & 13083 & Banking \\
\bottomrule
\end{tabular}
}
\caption{Dataset statistics.}
\label{tab:dataset_stats}
\end{table}

\subsection{Baselines and Experimental Settings}

This work defines Known Intent Ratio (KIR) to describe the class number of all intents that split into the known intent class from datasets, as follows:
\begin{equation}
\text{KIR}=\frac{|\mathcal{Y}_{\mathrm{ID}}|}{|\mathcal{Y}|},
\end{equation}
where $\mathcal{Y}$ is the full intent label set and $\mathcal{Y}_{\mathrm{ID}}$ is the subset of intents treated as known during training.

Moreover, the following state-of-the-art methods are selected as baselines: 

\textbf{MSP} \citep{hendrycks2017baseline} uses the maximum softmax probability as the confidence score for detecting misclassified or out-of-distribution samples.

\textbf{OpenMax} \citep{bendale2016towards} recalibrates activation scores and estimates the probability that an input belongs to an unknown class.

\textbf{DOC} \citep{shu2017doc} replaces the softmax layer with independent sigmoid classifiers and performs class-wise open-set rejection.

\textbf{DeepUnk} \citep{lin2019deep} uses margin loss to learn discriminative intent features and applies density-based novelty detection for unknown intent detection.

\textbf{KNNCL} \citep{zhou2022knn} combines KNN-based contrastive representation learning with density-based outlier detection for OOD intent classification.

\textbf{ADB} \citep{zhang2021adaptive} learns adaptive class-specific decision boundaries for open intent detection.

\textbf{DA-ADB} \citep{zhang2023learning} improves ADB with distance-aware representation learning and adaptive boundary learning.

Moreover, for the proposed cascade workflow, the model in gate module is 22M-parameter \texttt{all-MiniLM-L6-v2}. The models in router and expert modules are both 135M-parameter \texttt{SmolLM-135M}. The rank of LoRA (see section~\ref{sec: cascade}) is set as 32 for router, and set as 16 for expert. The scaling factor of LoRA is set as 64 for router, and set as 32 for expert. Due to the usage of LoRA, the parameter scaling is about 160M, which is close to that of BERT-base~\citep{zeng2021modeling, zhang2021adaptive, li2025multi}. The models are all trained on NVIDIA RTX 5070 GPU. The optimizer is AdamW with default settings. The batch size is 32. The learning rate is $2\times 10^{-4}$. The sub-centroid number for each intent $y$ in Eq.~(\ref{eq: ky}) is set as $K_y=2$. The boundary scaling factor in Eq.~(\ref{eq: lambda}) is set as $\lambda = 0.5$ for CLINC150 and $\lambda = 1$ for the other datasets. The embedding dimension in Eq.~(\ref{eq: embed}) is set as $h=384$.

\subsection{Evaluation Metrics}

To evaluate and compare the performances of all methods, Known F1 score, OOS F1 score and Acc are selected as metrics. 

\textbf{Known F1 score}~\citep{wang2021texsmart, zawbaa2024improved} is calculated by averaging F1 scores over all known intent classes, as follows:
\begin{equation}
\mathrm{Known\ F1}=\frac{1}{|\mathcal{Y}_{\mathrm{ID}}|}\sum_{y \in \mathcal{Y}_{\mathrm{ID}}}\frac{2P_yR_y}{P_y+R_y},
\end{equation}
where $\mathcal{Y}_{\mathrm{ID}}$ denotes the known-intent label set. $P_y$ and $R_y$ denote the precision and recall of class $y$, respectively.

\textbf{OOS F1 score}~\citep{wang2021texsmart, zawbaa2024improved} is calculated for evaluating OOS detection performance, as follows:
\begin{equation}
\mathrm{OOS\ F1}=\frac{2 P_{\mathrm{OOS}} R_{\mathrm{OOS}}}{P_{\mathrm{OOS}} + R_{\mathrm{OOS}}},
\end{equation}
where
\begin{equation}
\begin{aligned}
P_{\mathrm{OOS}}&=\frac{\mathrm{TP}_{\mathrm{OOS}}} {\mathrm{TP}_{\mathrm{OOS}}+\mathrm{FP}_{\mathrm{OOS}}}, \\
R_{\mathrm{OOS}}&=\frac{\mathrm{TP}_{\mathrm{OOS}}}
{\mathrm{TP}_{\mathrm{OOS}}+\mathrm{FN}_{\mathrm{OOS}}}.
\end{aligned}
\end{equation}
$\mathrm{TP}_{\mathrm{OOS}}$, $\mathrm{FP}_{\mathrm{OOS}}$ and $\mathrm{FN}_{\mathrm{OOS}}$ denote true positives, false positives and false negatives w.r.t. the OOS class, respectively.

\textbf{Acc} is the overall accuracy over all test samples, as follows:
\begin{equation}
\mathrm{Acc}=\frac{1}{N}\sum_{i=1}^{N}\mathbb{I}(\hat{y}_i = y_i),
\end{equation}
where $N$ is the size of test set, $y_i$ and $\hat{y}_i$ are the ground truth and the predicted labels of the $i$-th sample, respectively. And $\mathbb{I}(\cdot)$ is the indicator function.

\subsection{Performance on OOS Intent Detection}

The performance on OOS intent detection is presented in table~\ref{tab:main_results_all}. The results show that the proposed method achieves the state-of-the-art OOS F1 scores across all settings. Compared to the baselines, the proposed method achieves 0.85\%$\sim17.12\%$ improvements on OOS F1 scores.

As shown in table~\ref{tab:main_results_all}, the OOS F1 scores of all methods decrease as KIR increases. That means the dense known intent distributions intensify the semantic overlap between in-domain and OOS queries.

But the proposed method performs stability on OOS F1 scores under different KIRs. A possible reason is that our method maps user utterance into multiple clusters, that helps draw a clearer boundary compared to single cluster~\citep{zhang2021adaptive, zhang2023learning}. Moreover, the proposed method achieves outstanding performance on Banking77 dataset when KIR=0.25, but ranks second or third on known intent detection task. That means the supervised fine-tuning of MiniLMs used in router and experts yields unsatisfactory performance. But we think it still provides some insights for known intent detection task, though it is not the main contribution of this work. And the cascade workflow actually separates the OOS rejection and known intent detection process. Thus, if the other baseline methods can perform better than ours, it is easy to replace the router-experts pipeline with that method.

\begin{table*}[!t]
\centering
\tiny
\setlength{\tabcolsep}{2.8pt}
\renewcommand{\arraystretch}{0.97}
\resizebox{\textwidth}{!}{
\begin{tabular}{c|l|ccc|ccc|ccc}
\toprule
\multirow{2}{*}{\textbf{KIR}}
& \multirow{2}{*}{\textbf{Method}}
& \multicolumn{3}{c|}{\textbf{CLINC150}}
& \multicolumn{3}{c|}{\textbf{StackOverflow}}
& \multicolumn{3}{c}{\textbf{Banking77}} \\
\cmidrule{3-11}
& & \textbf{Known F1} & \textbf{OOS F1} & \textbf{Acc}
& \textbf{Known F1} & \textbf{OOS F1} & \textbf{Acc}
& \textbf{Known F1} & \textbf{OOS F1} & \textbf{Acc} \\
\midrule
\multirow{8}{*}{$0.25$}
& MSP      & 51.02 & 59.26 & 53.38 & 42.66 & 11.66 & 27.91 & 50.47 & 39.42 & 42.19 \\
& OpenMax  & 62.65 & 77.51 & 70.27 & 47.51 & 34.52 & 38.97 & 53.42 & 48.52 & 47.76 \\
& DOC      & 75.46 & 90.78 & 86.08 & 56.30 & 62.50 & 57.75 & 65.16 & 76.64 & 70.31 \\
& DeepUnk  & 76.95 & 91.61 & 87.18 & 47.39 & 36.87 & 40.03 & 64.97 & 76.98 & 70.68 \\
& KNNCL    & 78.85 & 93.56 & 89.87 & 41.79 & 15.26 & 28.65 & 65.54 & 79.34 & 73.01 \\
& ADB      & 77.85 & 92.36 & 88.30 & 77.62 & 90.96 & 86.75 & 70.92 & 85.05 & 79.33 \\
& DA-ADB   & \textbf{79.57} & 93.20 & 89.48 & \textbf{80.87} & 92.65 & 89.07 & 73.05 & 86.57 & 81.19 \\
& \textbf{Ours} & 71.75 & \textbf{95.01} & \textbf{90.45} & 73.61 & \textbf{94.47} & \textbf{91.04} & \textbf{75.83} & \textbf{93.99} & \textbf{89.07} \\
\midrule
\multirow{8}{*}{$0.50$}
& MSP      & 72.82 & 63.71 & 66.68 & 66.28 & 26.94 & 53.23 & 73.20 & 46.29 & 61.67 \\
& OpenMax  & 79.83 & 82.15 & 80.22 & 69.88 & 46.11 & 60.27 & 75.16 & 55.03 & 65.53 \\
& DOC      & 83.84 & 87.45 & 85.19 & 77.37 & 71.18 & 73.88 & 78.38 & 72.66 & 74.60 \\
& DeepUnk  & 83.30 & 87.48 & 84.95 & 67.67 & 35.80 & 55.46 & 75.61 & 67.80 & 71.01 \\
& KNNCL    & 83.25 & 87.85 & 85.32 & 61.50 & 8.50 & 45.38 & 75.16 & 67.21 & 70.41 \\
& ADB      & 85.12 & 88.60 & 86.54 & 85.32 & 87.70 & 86.49 & 81.39 & 79.43 & 79.61 \\
& DA-ADB   & \textbf{85.58} & 90.10 & \textbf{87.93} & \textbf{86.71} & 88.86 & \textbf{87.78} & \textbf{82.54} & 79.93 & \textbf{81.51} \\
& \textbf{Ours} & 79.95 & \textbf{91.96} & 86.78 & 75.48 & \textbf{89.71} & 85.54 & 74.90 & \textbf{88.23} & 78.98 \\
\midrule
\multirow{8}{*}{$0.75$}
& MSP      & 83.65 & 63.86 & 76.19 & 81.42 & 37.86 & 73.20 & 84.99 & 46.05 & 77.08 \\
& OpenMax  & 71.14 & 75.18 & 75.36 & 82.98 & 49.69 & 75.78 & 85.50 & 53.02 & 78.32 \\
& DOC      & 87.91 & 83.87 & 85.93 & 85.64 & 65.32 & 80.55 & 84.14 & 63.51 & 78.94 \\
& DeepUnk  & 86.57 & 82.67 & 84.61 & 80.51 & 34.38 & 71.56 & 81.65 & 50.57 & 74.73 \\
& KNNCL    & 86.14 & 82.05 & 84.12 & 76.16 & 7.19 & 65.01 & 81.76 & 51.42 & 74.78 \\
& ADB      & \textbf{88.97} & 84.85 & 86.99 & 86.91 & 74.10 & 82.89 & \textbf{86.44} & 67.34 & \textbf{81.39} \\
& DA-ADB   & 88.43 & 86.00 & \textbf{87.39} & \textbf{87.66} & 74.55 & 83.56 & 85.93 & 69.37 & 81.12 \\
& \textbf{Ours} & 66.32 & \textbf{87.10} & 79.83 & 80.18 & \textbf{75.57} & 81.32 & 70.28 & \textbf{86.49} & 77.84 \\
\bottomrule
\end{tabular}
}
\caption{Performance comparison for all methods under different KIRs.}
\label{tab:main_results_all}
\end{table*}

\subsection{Interpretability Analysis for Gate Stage}

\begin{figure}[t]
    \centering
    \includegraphics[width=0.9\linewidth]{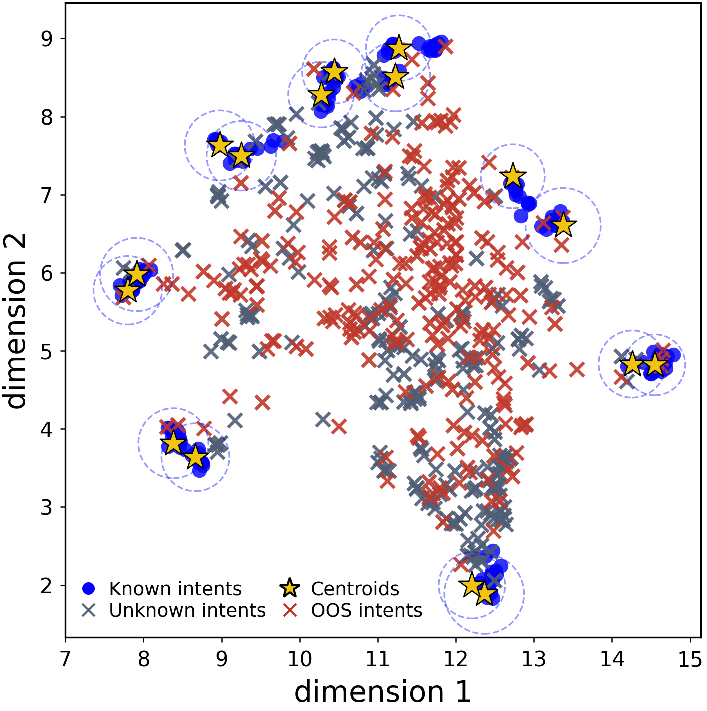}
    \caption{An example of MiniLM embeddings from CLINC150 dataset when $\mathrm{KIR}=0.50$.}
    \label{fig:gate_embedding}
\end{figure}

\begin{figure}[t]
    \centering
    \includegraphics[width=\columnwidth]{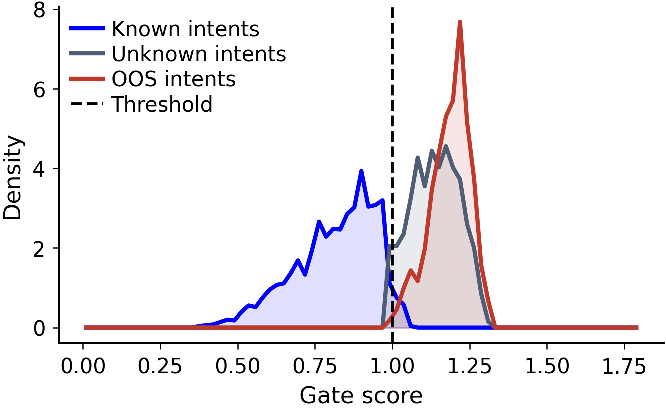}
    \caption{Gate score distribution on CLINC150 under $\mathrm{KIR}=0.50$. The dashed line marks $s(\bar{e})=1$.}
    \label{fig:gate_score_distribution}
\end{figure}

Figure~\ref{fig:gate_embedding} provides an example, that shows the MiniLM embedding results on two dimensions. It is clear that the embeddings of known intents generate multiple clusters, and the utterance embedding with the same intent class would concentrate into the same cluster. This makes it easy to discover the boundary for all known intents, especially with higher dimension, and the outside space is out-of-scope. A clearer evidence to show the separation is presented in figure~\ref{fig:gate_score_distribution}. According to our method, the threshold score strictly separates the two distributions of the known intent samples and OOS intent samples. 

Moreover, some OOS intent samples are used to help learn the parameter $\lambda$ in Eq.~(\ref{eq: lambda}). They are viewed as unknown intent samples in the validation set. But actually, although there is a bit of overlap between samples of the unknown intent samples and the OOS intent test samples as shown in figure~\ref{fig:gate_embedding}, the distributions of them are different, as shown in figure~\ref{fig:gate_score_distribution}. That means, the distribution of the unknown intent samples do not influence the learning performances. 

\subsection{MiniLM Selection for Gate stage}

The ablation studies for MiniLM selection of gate stage are also conducted, as shown in table~\ref{tab:ablation_all_kir}. In the table, \textbf{Without Gate} means to cancel the gate stage in our full pipeline, and instead conduct OOS rejection in the router module. \textbf{Cascade-MiniLM} and \textbf{Cascade-SmolLM} means to use the MiniLMs in all stages. 

\begin{table*}[t]
  \centering
  \small
  \setlength{\tabcolsep}{0pt}
  \renewcommand{\arraystretch}{1.}
  \begin{tabular*}{\textwidth}{@{\extracolsep{\fill}}clcccccc@{}}
  \toprule
  \multirow{2}{*}{\textbf{KIR}}
  & \multirow{2}{*}{\textbf{Variant}}
  & \multicolumn{2}{c}{\textbf{CLINC150}}
  & \multicolumn{2}{c}{\textbf{StackOverflow}}
  & \multicolumn{2}{c}{\textbf{Banking77}} \\
  \cmidrule(lr){3-4}
  \cmidrule(lr){5-6}
  \cmidrule(lr){7-8}
  & & \textbf{Acc} & \textbf{OOS F1}
  & \textbf{Acc} & \textbf{OOS F1}
  & \textbf{Acc} & \textbf{OOS F1} \\
  \midrule
  \multirow{4}{*}{$0.25$}
  & \textbf{Ours}  & \textbf{90.45} & \textbf{95.01} & \textbf{91.04} & \textbf{94.47} & \textbf{89.07} & \textbf{93.99} \\
  & Without Gate   & 67.49 & 76.88 & 81.80 & 88.30 & 84.90 & 91.27 \\
  & Cascade-MiniLM & 90.07 & 94.14 & 79.73 & 86.85 & 85.10 & 91.41 \\
  & Cascade-SmolLM & 84.76 & 90.98 & 40.28 & 45.00 & 51.20 & 64.43 \\
  \midrule
  \multirow{4}{*}{$0.50$}
  & \textbf{Ours}  & \textbf{86.78} & \textbf{91.96} & \textbf{85.54} & \textbf{89.71} & \textbf{78.98} & \textbf{88.23} \\
  & Without Gate   & 67.90 & 70.24 & 80.28 & 82.79 & 77.15 & 82.88 \\
  & Cascade-MiniLM & 85.89 & 90.67 & 77.46 & 79.41 & 77.11 & 84.63 \\
  & Cascade-SmolLM & 73.58 & 78.69 & 50.10 & 55.68 & 65.83 & 77.61 \\
  \midrule
  \multirow{4}{*}{$0.75$}
  & \textbf{Ours}  & \textbf{79.83} & \textbf{87.10} & \textbf{81.32} & \textbf{75.57} & \textbf{77.84} & \textbf{86.49} \\
  & Without Gate   & 73.40 & 66.50 & 77.36 & 65.83 & 76.50 & 84.20 \\
  & Cascade-MiniLM & 79.10 & 81.00 & 75.91 & 62.50 & 77.23 & 83.69 \\
  & Cascade-SmolLM & 74.31 & 71.79 & 58.91 & 28.72 & 52.48 & 55.41 \\
  \bottomrule
  \end{tabular*}
  \caption{Ablation results under different KIR settings.}
  \label{tab:ablation_all_kir}
\end{table*}

The results that the full pipeline of ours achieves the state-of-the-art performance, compared the ablation variants. That means, (i) the gate stage is essential for OOS rejection, (ii) hybrid MiniLMs in ours have more advantages.

Moreover, we first use the 135M-parameter \texttt{SmolLM-135M} for all stages actually. But after that, we replace it with the 22M-parameter \texttt{all-MiniLM-L6-v2} for gate stage, and the performance is improved. That means, it is not essential to use models with large parameters, which refers to our insight on LLMs in section Related Work.

\section{Conclusion}

This work proposes a multi-cluster boundary learning method for OOS intent detection via a cascade MiniLM workflow. That workflow separates the OOS intent detection from the multi-class classification of known intents, and views it as a one-class classification problem. Then, multi-cluster boundary learning is conducted to learn the boundary of MiniLM embeddings from the input user utterances. That boundary separates the embedding domain of OOS intents and known intents. Extensive experiments are conducted on public datasets. The results show that the proposed method achieve the stete-of-the-art OOS intent detection performance. Ablation studies also show that the MiniLM settings are reasonable at present. Moreover, the performance of the method is not always the best one on the task of known intent detection, but the cascade workflow completely decouples the two detection process for OOS and known intents. Thus, the result does not influence the aforementioned conclusion. And moreover, the lightweight models have outstanding advantages for real-world applications compared to the large models. The more advanced MiniLMs perhaps improve the detection performance by using this framework in the future.

%

\bibliography{custom}

@inproceedings{larson2019evaluation,
  title = {An Evaluation Dataset for Intent Classification and Out-of-Scope Prediction},
  author = {Larson, Stefan and Mahendran, Anish and Peper, Joseph J. and Clarke, Christopher and Lee, Andrew and Hill, Parker and Kummerfeld, Jonathan K. and Leach, Kevin and Laurenzano, Michael A. and Tang, Lingjia and Mars, Jason},
  booktitle = {EMNLP},
  year = {2019},
  note = {\url{https://aclanthology.org/D19-1131/}}
}

@inproceedings{louvan2020recent,
  title = {Recent Neural Methods on Slot Filling and Intent Classification for Task-Oriented Dialogue Systems: A Survey},
  author = {Louvan, Samuel and Magnini, Bernardo},
  booktitle = {COLING},
  pages = {480-496},
  year = {2020},
  note = {\url{https://aclanthology.org/2020.coling-main.42/}}
}

@inproceedings{wang2002combination,
  title = {Combination of Statistical and Rule-Based Approaches for Spoken Language Understanding},
  author = {Wang, Ye-Yi and Acero, Alex and Chelba, Ciprian and Frey, Brendan and Wong, Leon},
  booktitle = {ICSLP},
  pages = {609--612},
  year = {2002},
  url = {https://doi.org/10.21437/ICSLP.2002-204}
}

@inproceedings{tur2010what,
  title = {What is Left to be Understood in {ATIS}?},
  author = {Tur, Gokhan and Hakkani-T{\"u}r, Dilek and Heck, Larry},
  booktitle = {IEEE Spoken Language Technology Workshop},
  year = {2010},
  note = {\url{https://doi.org/10.1109/SLT.2010.5700816}}
}

@inproceedings{haffner2003optimizing,
  title = {Optimizing {SVM}s for Complex Call Classification},
  author = {Haffner, Patrick and Tur, Gokhan and Wright, Jerome H.},
  booktitle = {ICASSP},
  year = {2003},
  note = {\url{https://doi.org/10.1109/ICASSP.2003.1198860}}
}

@article{schuurmans2020intent,
  title = {Intent Classification for Dialogue Utterances},
  author = {Schuurmans, Jetze and Frasincar, Flavius},
  journal = {IEEE Intelligent Systems},
  volume = {35},
  number = {1},
  pages = {82-88},
  year = {2020},
  note = {\url{https://doi.org/10.1109/MIS.2019.2954966}}
}

@inproceedings{hendrycks2017baseline,
  title = {A Baseline for Detecting Misclassified and Out-of-Distribution Examples in Neural Networks},
  author = {Hendrycks, Dan and Gimpel, Kevin},
  booktitle = {ICLR},
  year = {2017},
  note = {\url{https://openreview.net/forum?id=Hkg4TI9xl}}
}

@inproceedings{guo2017calibration,
  title = {On Calibration of Modern Neural Networks},
  author = {Guo, Chuan and Pleiss, Geoff and Sun, Yu and Weinberger, Kilian Q.},
  booktitle = {ICML},
  year = {2017},
  note = {\url{https://proceedings.mlr.press/v70/guo17a.html}}
}

@inproceedings{bendale2016towards,
  title = {Towards Open Set Deep Networks},
  author = {Bendale, Abhijit and Boult, Terrance E.},
  booktitle = {CVPR},
  year = {2016},
  note = {\url{https://openaccess.thecvf.com/content_cvpr_2016/html/Bendale_Towards_Open_Set_CVPR_2016_paper.html}}
}

@inproceedings{shu2017doc,
  title = {{DOC}: Deep Open Classification of Text Documents},
  author = {Shu, Lei and Xu, Hu and Liu, Bing},
  booktitle = {EMNLP},
  year = {2017},
  note = {\url{https://aclanthology.org/D17-1314/}}
}

@inproceedings{lin2019deep,
  title = {Deep Unknown Intent Detection with Margin Loss},
  author = {Lin, Ting-En and Xu, Hua},
  booktitle = {ACL},
  year = {2019},
  note = {\url{https://aclanthology.org/P19-1548/}}
}

@inproceedings{zhou2022knn,
  title = {{KNN}-Contrastive Learning for Out-of-Domain Intent Classification},
  author = {Zhou, Yunhua and Liu, Peiju and Qiu, Xipeng},
  booktitle = {ACL},
  year = {2022},
  note = {\url{https://aclanthology.org/2022.acl-long.352/}}
}

@inproceedings{lee2018simple,
  title = {A Simple Unified Framework for Detecting Out-of-Distribution Samples and Adversarial Attacks},
  author = {Lee, Kimin and Lee, Kibok and Lee, Honglak and Shin, Jinwoo},
  booktitle = {NeurIPS},
  year = {2018},
  note = {\url{https://papers.nips.cc/paper_files/paper/2018/hash/abdeb6f575ac5c6676b747bca8d09cc2-Abstract.html}}
}

@inproceedings{zeng2021modeling,
  title = {Modeling Discriminative Representations for Out-of-Domain Detection with Supervised Contrastive Learning},
  author = {Zeng, Zhiyuan and He, Keqing and Yan, Yuanmeng and Liu, Zijun and Wu, Yanan and Xu, Hong and Jiang, Huixing and Xu, Weiran},
  booktitle = {ACL and IJCNLP},
  year = {2021},
  note = {\url{https://aclanthology.org/2021.acl-short.110/}}
}

@inproceedings{zhang2021adaptive,
  title = {Deep Open Intent Classification with Adaptive Decision Boundary},
  author = {Zhang, Hanlei and Xu, Hua and Lin, Ting-En},
  booktitle = {AAAI},
  year = {2021},
  note = {\url{https://ojs.aaai.org/index.php/AAAI/article/view/17690}}
}

@article{zhang2023learning,
  title = {Learning Discriminative Representations and Decision Boundaries for Open Intent Detection},
  author = {Zhang, Hanlei and Xu, Hua and Zhao, Shaojie and Zhou, Qianrui},
  journal = {IEEE/ACM Transactions on Audio, Speech, and Language Processing},
  volume = {31},
  pages={1611-1623},
  year = {2023},
  note = {\url{https://ieeexplore.ieee.org/document/10102404}}
}

@inproceedings{li2025multi,
  title = {Multi-Granularity Open Intent Classification via Adaptive Granular-Ball Decision Boundary},
  author = {Li, Yanhua and Ouyang, Xiaocao and Pan, Chaofan and Zhang, Jie and Zhao, Sen and Xia, Shuyin and Yang, Xin and Wang, Guoyin and Li, Tianrui},
  booktitle = {AAAI},
  year = {2025},
  note = {\url{https://ojs.aaai.org/index.php/AAAI/article/view/34630}}
}

@inproceedings{zaera2025efficient,
  title = {Efficient Out-of-Scope Detection in Dialogue Systems via Uncertainty-Driven {LLM} Routing},
  author = {Zaera, {\'A}lvaro and Popa, Diana Nicoleta and Sekulic, Ivan and Rosso, Paolo},
  booktitle = {ACL},
  year = {2025},
  note = {\url{https://aclanthology.org/2025.acl-industry.25/}}
}

@article{hoffman2024inferring,
  title = {Inferring Human Intent and Predicting Human Action in Human-Robot Collaboration},
  author = {Hoffman, Guy and Bhattacharjee, Tapomayukh and Nikolaidis, Stefanos},
  journal = {Annual Review of Control, Robotics, and Autonomous Systems},
  volume = {7},
  pages = {73-95},
  year = {2024},
  note = {\url{https://doi.org/10.1146/annurev-control-071223-105834}}
}

@article{muzahid2024survey,
  title = {Survey on Human-Vehicle Interactions and {AI} Collaboration for Optimal Decision-Making in Automated Driving},
  author = {Muzahid, Abu Jafar Md and Zhao, Xiaopeng and Wang, Zhenbo},
  journal = {arXiv},
  year = {2024},
  note = {\url{https://arxiv.org/abs/2412.08005}}
}

@inproceedings{wolflein2025agents,
  title = {{LLM} Agents Making Agent Tools},
  author = {W{\"o}lflein, Georg and Ferber, Dyke and Truhn, Daniel and Arandjelovi{\'c}, Ognjen and Kather, Jakob Nikolas},
  booktitle = {ACL},
  year = {2025},
  note = {\url{https://aclanthology.org/2025.acl-long.1266/}}
}

@inproceedings{arora2024intent,
  title = {Intent Detection in the Age of {LLM}s},
  author = {Arora, Gaurav and Jain, Shreya and Merugu, Srujana},
  booktitle = {EMNLP},
  year = {2024},
  note = {\url{https://aclanthology.org/2024.emnlp-industry.114/}}
}

@inproceedings{kim2014convolutional,
  title = {Convolutional Neural Networks for Sentence Classification},
  author = {Kim, Yoon},
  booktitle = {EMNLP},
  year = {2014},
  note = {\url{https://aclanthology.org/D14-1181/}}
}

@inproceedings{liu2016attention,
  title = {Attention-Based Recurrent Neural Network Models for Joint Intent Detection and Slot Filling},
  author = {Liu, Bing and Lane, Ian},
  booktitle = {INTERSPEECH},
  year = {2016},
  note = {\url{https://www.isca-archive.org/interspeech_2016/liu16c_interspeech.html}}
}

@inproceedings{vaswani2017attention,
  title = {Attention Is All You Need},
  author = {Vaswani, Ashish and Shazeer, Noam and Parmar, Niki and Uszkoreit, Jakob and Jones, Llion and Gomez, Aidan N. and Kaiser, {\L}ukasz and Polosukhin, Illia},
  booktitle = {NeurIPS},
  year = {2017},
  note = {\url{https://proceedings.neurips.cc/paper/2017/hash/3f5ee243547dee91fbd053c1c4a845aa-Abstract.html}}
}

@inproceedings{chen2025collaborating,
  title = {On Collaborating Small and Large Models For Few-shot Intent Detection},
  author = {Chen, Peng and Wang, Bang},
  booktitle = {EMNLP Findings},
  year = {2025},
  note = {\url{https://aclanthology.org/2025.findings-emnlp.749/}}
}

@inproceedings{hu2022lora,
  title = {LoRA: Low-Rank Adaptation of Large Language Models},
  author = {Hu, Edward J. and Shen, Yelong and Wallis, Phillip and Allen-Zhu, Zeyuan and Li, Yuanzhi and Wang, Shean and Wang, Lu and Chen, Weizhu},
  booktitle = {ICLR},
  year = {2022},
  note = {\url{https://openreview.net/forum?id=nZeVKeeFYf9}}
}

@inproceedings{zawbaa2024improved,
  title = {Improved Out-of-Scope Intent Classification with Dual Encoding and Threshold-based Re-Classification},
  author = {Zawbaa, Hossam M. and Rashwan, Wael and Dutta, Sourav and Assem, Haytham},
  booktitle = {LREC and COLING},
  year = {2024},
  note = {\url{https://aclanthology.org/2024.lrec-main.763/}}
}

@inproceedings{wang2021texsmart,
  title = {TexSmart: A System for Enhanced Natural Language Understanding},
  author = {Liu, Lemao and Zhang, Haisong and Jiang, Haiyun and Li, Yangming and Zhao, Enbo and Xu, Kun and Song, Linfeng and Zheng, Suncong and Zhou, Botong and Zhu, Dick and Feng, Xiao and Chen, Tao and Yang, Tao and Yu, Dong and Zhang, Feng and Kang, Zhanhui and Shi, Shuming},
  booktitle = {ACL Demo},
  year = {2021},
  note = {\url{https://aclanthology.org/2021.acl-demo.20/}}
}

\end{document}